\documentclass[10pt,twocolumn,letterpaper]{article}
\usepackage{cvpr}
\usepackage{times}
\usepackage{graphicx}
\usepackage{amsmath}
\usepackage{amssymb}
\usepackage{xspace}
\usepackage{xcolor}
\usepackage{enumitem}
\usepackage{listings}
\usepackage{tabularx}
\usepackage{multirow}
\usepackage[export]{adjustbox}    
\usepackage{tikz}
\usepackage{pgffor}
\usepackage[numbers,square,sort&compress]{natbib}
\usepackage[
  pagebackref=true,
  breaklinks=true,
  letterpaper=true,
  colorlinks,
  bookmarks=false]{hyperref}

\cvprfinalcopy
\def\cvprPaperID{532}

\graphicspath{{./figures}{./}{./figures/v14}}

\setitemize{noitemsep,topsep=0pt,parsep=0pt,partopsep=0pt,nolistsep,leftmargin=*}

\renewcommand{\paragraph}[1]{\medskip\noindent{\bf #1}}

\definecolor{attr}{rgb}{0.85,0.88,0.90}
\definecolor{code}{rgb}{0.95,0.93,0.90}

\newcommand{\off}[1]{}

\newcommand{\bx}{\mathbf{x}}

\newcommand{\bu}{\mathbf{u}}
\newcommand{\bg}{\mathbf{g}}

\newcommand{\real}{\mathbb{R}}

\makeatletter
\g@addto@macro\normalsize{%
  \setlength\abovedisplayskip{0.7em}
  \setlength\belowdisplayskip{0.7em}
  \setlength\abovedisplayshortskip{0.7em}
  \setlength\belowdisplayshortskip{0.7em}
}
\makeatother

\tikzset{
 image label/.style={
   fill=white,
   text=black,
   font=\tiny,
   anchor=south east,
   xshift=-0.1cm,
   yshift=0.1cm,
   at={(0,0)}
 }
}
 
\begin{document}
\title{Understanding Deep Image Representations by Inverting Them}
\vspace{-2em}
\author{
Aravindh Mahendran \\
University of Oxford
\and
Andrea Vedaldi \\
University of Oxford}
\maketitle

\begin{abstract}
Image representations, from SIFT and Bag of Visual Words to Convolutional Neural Networks (CNNs), are a crucial component of almost any image understanding system.
Nevertheless, our understanding of them remains limited.
In this paper we conduct a direct analysis of the visual information contained in representations by asking the following question: given an encoding of an image, to which extent is it possible to reconstruct the image itself?
To answer this question we contribute a general framework to invert representations.
We show that this method can invert representations such as HOG and SIFT more accurately than recent alternatives while being applicable to CNNs too.
We then use this technique to study the inverse of recent state-of-the-art CNN image representations for the first time.
Among our findings, we show that several layers in CNNs retain photographically accurate information about the image, with different degrees of geometric and photometric invariance.
\end{abstract}
\section{Introduction}\label{s:intro}

Most image understanding and computer vision methods build on  image representations such as textons~\cite{leung01representing}, histogram of oriented gradients (SIFT~\cite{lowe04distinctive} and HOG~\cite{dalal05histograms}), bag of visual words~\cite{csurka04visual}\cite{sivic03video}, sparse~\cite{yang10supervised} and local coding~\cite{wang10locality-constrained}, super vector coding~\cite{zhou10image}, VLAD~\cite{jegou10aggregating}, Fisher Vectors~\cite{perronnin06fisher}, and, lately, deep neural networks, particularly of the convolutional variety~\cite{krizhevsky12imagenet,zeiler14visualizing,sermanet14overfeat:}.
However, despite the progress in the development of visual representations, their design is still driven empirically and a good understanding of their properties is lacking.
While this is true of shallower hand-crafted features, it is even more so for the latest generation of deep representations, where millions of parameters are learned from data.

\begin{figure}
\begin{tikzpicture}
\node[anchor=north west,inner sep=0] at (0,0) {
\includegraphics[width=\columnwidth]{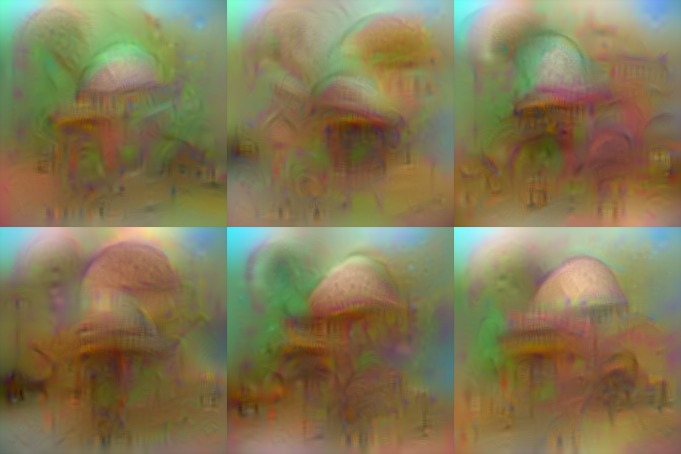}
};
\node[anchor=north west,inner sep=0,at={(0,0)}] (I) {
\includegraphics[width=0.3333\columnwidth]{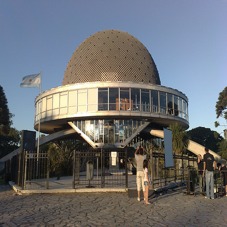}
};
\end{tikzpicture}
\caption{{\bf What is encoded by a CNN?} The figure shows five possible reconstructions of the reference image obtained from the 1,000-dimensional code extracted at the penultimate layer of a reference CNN\cite{krizhevsky12imagenet} (before the softmax is applied) trained on the ImageNet data. From the viewpoint of the model, all these images are practically equivalent. This image is best viewed in color/screen.}\label{f:intro}
\end{figure}

In this paper we conduct a direct analysis of representations by characterising the image information that they retain (Fig.~\ref{f:intro}).
We do so by modeling a representation as a function $\Phi(\bx)$ of the image $\bx$ and then computing an approximated inverse $\phi^{-1}$, \emph{reconstructing $\bx$ from the code $\Phi(\bx)$}.
A common hypothesis is that representations collapse irrelevant differences in images (e.g. illumination or viewpoint), so that $\Phi$ should not be uniquely invertible.
Hence, we pose this as a reconstruction problem and find a number of possible reconstructions rather than a single one.
By doing so, we obtain insights into the invariances captured by the representation.

Our contributions are as follows.
First, we propose a general method to invert representations, including SIFT, HOG, and CNNs (Sect.~\ref{s:method}).
Crucially, this method {\bf uses only information from the image representation} and a generic natural image prior, starting from random noise as initial solution, and hence captures only the information contained in the representation itself.
We discuss and evaluate different regularization penalties as natural image priors.
Second, we show that, despite its simplicity and generality, this method recovers significantly better reconstructions from DSIFT and HOG compared to recent alternatives~\cite{vondrick13hoggles:}.
As we do so, we emphasise a number of subtle differences between these representations and their effect on invertibility.
Third, we apply the inversion technique to the analysis of recent deep CNNs, exploring their invariance by sampling possible approximate reconstructions.
We relate this to the depth of the representation, showing that the CNN gradually builds an increasing amount of invariance, layer after layer.
Fourth, we study the locality of the information stored in the representations by reconstructing images from selected groups of neurons, either spatially or by channel.

The rest of the paper is organised as follows.
Sect.~\ref{s:method} introduces the inversion method, posing this as a regularised regression problem and proposing a number of image priors to aid the reconstruction.
Sect.~\ref{s:representations} introduces various representations: HOG and DSIFT as examples of shallow representations, and state-of-the-art CNNs as an example of deep representations.
It also shows how HOG and DSIFT can be implemented as CNNs, simplifying the computation of their derivatives. Sect.~\ref{s:results-shallow} and~\ref{s:results-deep} apply the inversion technique to the analysis of respectively shallow (HOG and DSIFT) and deep (CNNs) representations.
Finally, Sect.~\ref{s:summary} summarises our findings.

We use the matconvnet toolbox~\cite{matconvnet2014vedaldi} for implementing convolutional neural networks.

\paragraph{Related work.} There is a significant amount of work in understanding representations by means of visualisations.
The works most related to ours are Weinzaepfel~\etal~\cite{weinzaepfel11reconstructing} and Vondrick~\etal~\cite{vondrick13hoggles:} which invert sparse DSIFT and HOG features respectively.
While our goal is similar to theirs, our method is substantially different from a technical viewpoint, being based on the direct solution of a regularised regression problem.
The benefit is that our technique applies equally to shallow (SIFT, HOG) and deep (CNN) representations. Compared to existing inversion techniques for dense shallow representations~\cite{vondrick13hoggles:}, it is also shown to achieve superior results, both quantitatively and qualitatively.

An interesting conclusion of~\cite{weinzaepfel11reconstructing,vondrick13hoggles:} is that, while HOG and SIFT may not be exactly invertible, they capture a significant amount of information about the image. 
This is in apparent contradiction with the results of Tatu~\etal~\cite{tatu11exploring} who show that it is possible to make any two images look nearly identical in SIFT space up to the injection of  adversarial noise.
A symmetric effect was demonstrated for CNNs by Szegedy~\etal~\cite{szegedy13intriguing}, where an imperceptible amount of adversarial noise suffices to change the predicted class of an image.
The apparent inconsistency is easily resolved, however, as the methods of~\cite{tatu11exploring,szegedy13intriguing} require the injection of high-pass structured noise which is very unlikely to occur in natural images.

Our work is also related to the DeConvNet method of Zeiler and Fergus~\cite{zeiler14visualizing}, who backtrack the network computations to identify which image patches are responsible for certain neural activations.
Simonyan~\etal~\cite{simonyan14deep}, however, demonstrated that DeConvNets can be interpreted as a sensitivity analysis of the network input/output relation.
A consequence is that  DeConvNets do not study the problem of representation inversion in the sense adopted here, which has significant methodological consequences; for example, DeConvNets require \emph{auxiliary information} about the activations in several intermediate layers, while our inversion uses only the final image code.
In other words, DeConvNets look at \emph{how} certain network outputs are obtained, whereas we look for \emph{what} information is preserved by the network output.

The problem of inverting representations, particularly CNN-based ones, is related to the problem of inverting neural networks, which received significant attention in the past.
Algorithms similar to the back-propagation technique developed here were proposed by~\cite{williams86inverting,linden89inversion,lee94inverse,lu99inverting}, along with alternative optimisation strategies based on sampling.
However, these methods did not use natural image priors as we do, nor were applied to the current generation of deep networks.
Other works~\cite{jensen99inversion,varkonyi-koczy05observer} specialised on inverting networks in the context of dynamical systems and will not be discussed further here.
Others~\cite{bishop95neural} proposed to learn a second neural network to act as the inverse of the original one, but this is complicated by the fact that the inverse is usually not unique.
Finally, auto-encoder architectures~\cite{hinton06reducing} train networks together with their inverses as a form of supervision; here we are interested instead in visualising feed-forward and discriminatively-trained CNNs now popular in computer vision.

\section{Inverting representations}\label{s:method}

This section introduces our method to compute an approximate inverse of an image representation.
This is formulated as the problem of finding an image whose representation best matches the one given~\cite{williams86inverting}.
Formally, given a representation function $\Phi : \real^{H\times W \times C} \rightarrow \real^d$ and a representation $\Phi_0 = \Phi(\bx_0)$ to be inverted, reconstruction finds the image $\bx\in\real^{H \times W \times C}$ that minimizes the objective:
\begin{equation}\label{e:objective}
 \bx^* = \operatornamewithlimits{argmin}_{\bx\in\real^{H \times W \times C}} \ell(\Phi(\bx), \Phi_0) + \lambda \mathcal{R}(\bx)
\end{equation}
where the loss $\ell$ compares the image representation $\Phi(\bx)$ to the target one $\Phi_0$ and $\mathcal{R} : \real^{H \times W \times C} \rightarrow \real$ is a regulariser capturing a \emph{natural image prior}.

Minimising \eqref{e:objective} results in an image $\bx^*$ that ``resembles'' $\bx_0$ from the viewpoint of the representation.
While there may be no unique solution to this problem, sampling the space of possible reconstructions can be used to characterise the space of images that the representation deems to be equivalent, revealing its invariances.

We next discusses the choice of loss and regularizer.

\paragraph{Loss function.} There are many possible choices of the loss function $\ell$.
While we use the Euclidean distance:
\begin{equation}\label{e:objective2}
 \ell(\Phi(\bx),\Phi_0) = \| \Phi(\bx) - \Phi_0 \|^2,
\end{equation}
it is possible to change the nature of the loss entirely, for example to optimize selected neural responses.
The latter was used in~\cite{erhan09visualizing,simonyan14deep} to generate images representative of given neurons.

\paragraph{Regularisers.} Discriminatively-trained representations may discard a significant amount of low-level image statistics as these are usually not interesting for high-level tasks.
As this information is nonetheless useful for visualization, it can be partially recovered by restricting the inversion to the subset of natural images $\mathcal{X}\subset \real^{H\times W \times C}$.
However, minimising over $\mathcal{X}$ requires addressing the challenge of modeling this set.
As a proxy one can incorporate in the reconstruction an appropriate \emph{image prior}.
Here we experiment with two such priors.
The first one is simply the $\alpha$-norm  $\mathcal{R}_\alpha(\bx) = \|\bx\|_\alpha^\alpha$, where $\bx$ is the vectorised and mean-subtracted image.
By choosing a relatively large exponent ($\alpha=6$ is used in the experiments) the range of the image is encouraged to stay within a target interval instead of diverging.

\newcommand{\TV}{{V^\beta}}

\begin{figure}
\hfill
{\adjincludegraphics[width=0.33\columnwidth,trim={120pt 0 0 120pt},clip]{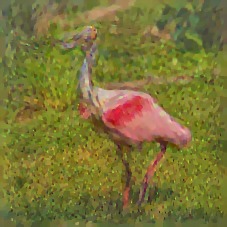}}
\hfill
{\adjincludegraphics[width=0.33\columnwidth,trim={120pt 0 0 120pt},clip]{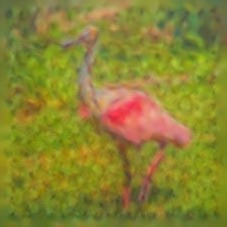}}
\hfill\mbox{}
\label{fig:spikes}
\caption{\textbf{Left:} Spikes in a inverse of norm1 features - detail shown. \textbf{Right:} Spikes removed by a $\TV$ regularizer with $\beta = 2$.}
\end{figure}

A second richer regulariser is \emph{total variation} (TV) $\mathcal{R}_\TV(\bx)$, encouraging images to consist of piece-wise constant patches.
For continuous functions (or distributions) $f : \real^{H\times W} \supset \Omega \rightarrow \real$, the TV norm is given by:
\[
 \mathcal{R}_\TV(f)
 =
 \int_{\Omega} 
 \left(
 \left(\frac{\partial f}{\partial u}(u,v)\right)^2 + 
 \left(\frac{\partial f}{\partial v}(u,v)\right)^2
 \right)^{\frac{\beta}{2}}\,du\,dv
\]
where $\beta = 1$.
Here images are discrete ($\bx \in \real^{H \times W}$) and the TV norm is replaced by the finite-difference approximation:
\[
 \mathcal{R}_\TV(\bx)
 =
 \sum_{i,j}
 \left(
 \left(x_{i,j+1} - x_{ij}\right)^2 +
 \left(x_{i+1,j} - x_{ij}\right)^2
 \right)^\frac{\beta}{2}.
\] 
It was observed empirically that the TV regularizer ($\beta = 1$) in the presence of subsampling, also caused by max pooling in CNNs, leads to ``spikes'' in the reconstruction.
This is a known problem in TV-based image interpolation (see \eg Fig.~3 in \cite{chen2014bi}) and is illustrated in Fig.~\ref{fig:spikes}.left when inverting a layer in a CNN.
The ``spikes'' occur at the locations of the samples because: (1) the TV norm along any path between two samples depends only on the overall amount of intensity change (not on the sharpness of the changes) and (2) integrated on the 2D image, it is optimal to concentrate sharp changes around a boundary with a small perimeter.
Hyper-Laplacian priors with $\beta < 1$ are often used as a better match of the gradient statistics of natural images~\cite{krishnan09fast}, but they only exacerbate this issue.
Instead, we trade-off the sharpness of the image with the removal of such artifacts by choosing $\beta > 1$ which, by penalising large gradients, distributes changes across regions rather than concentrating them at a point or curve.
We refer to this as the $\TV$ regularizer.
As seen in Fig.~\ref{fig:spikes} (right), the spikes are removed with $\beta = 2$ but the image is washed out as edges are penalized more than with $\beta = 1$.

When the target of the reconstruction is a colour image, both regularisers are summed for each colour channel.

\paragraph{Balancing the different terms.} Balancing loss and regulariser(s) requires some attention.
While an optimal tuning can be achieved by cross-validation, it is important to start from reasonable settings of the parameters.
First, the loss is replaced by the normalized version $\|\Phi(\bx) - \Phi_0\|^2_2/\|\Phi_0\|^2_2$.
This fixes its dynamic range, as after normalisation the loss near the optimum can be expected to be contained in the $[0,1)$ interval, touching zero at the optimum.
In order to make the dynamic range of the regulariser(s) comparable one can aim for a solution $\bx^*$ which has roughly unitary Euclidean norm.
While representations are largely insensitive to the scaling of the image range, this is not exactly true for the first few layers of CNNs, where biases are tuned to a ``natural'' working range.
This can be addressed by considering the objective $\|\Phi(\sigma \bx) - \Phi_0\|^2_2/\|\Phi_0\|^2_2 + \mathcal{R}(\bx)$ where the scaling $\sigma$ is the average Euclidean norm of natural images in a training set.

Second, the multiplier $\lambda_\alpha$ of the $\alpha$-norm regularizer should be selected to encourage the reconstructed image $\sigma\bx$ to be contained in a natural range $[-B, B]$ (\eg in most CNN implementations $B=128$).
If most pixels in $\sigma
\bx$ have a magnitude similar to $B$, then $\mathcal{R}_\alpha(\bx)\approx HWB^\alpha/\sigma^\alpha$, and $\lambda_\alpha\approx \sigma^\alpha/(HWB^\alpha)$.
A similar argument suggests to pick the $\TV$-norm regulariser coefficient as $\lambda_{\TV} \approx \sigma^\beta/(HW(aB)^\beta)$, where $a$ is a small fraction (\eg $a=1\%$) relating the dynamic range of the image to that of its gradient.

The final form of the objective function is 
\begin{equation}
\|\Phi(\sigma \bx) - \Phi_0\|^2_2/\|\Phi_0\|^2_2 + \lambda_\alpha \mathcal{R}_\alpha(\bx) + \lambda_\TV \mathcal{R}_\TV(\bx) \label{eq:obj}
\end{equation}
It is in general non convex because of the nature of $\Phi$. We next discuss how to optimize it.

\subsection{Optimisation}\label{s:optimisation}

Finding an optimizer of the objective~\eqref{e:objective} may seem a hopeless task as most representations $\Phi$ involve strong non-linearities; in particular, deep representations are a chain of \emph{several non-linear layers}.
Nevertheless, simple gradient descent (GD) procedures have been shown to be very effective in \emph{learning} such models from data, which is arguably an even harder task.
Hence, it is not unreasonable to use GD to solve~\eqref{e:objective} too.
We extend GD to incorporate a few extensions that proved useful in learning deep networks~\cite{krizhevsky12imagenet}, as discussed below.

\paragraph{Momentum.} GD is extended to use \emph{momentum}:
\[
 \mu_{t+1} \leftarrow m \mu_{t} - \eta_t \nabla E(\bx),\qquad
 \bx_{t+1} \leftarrow \bx_{t} + \mathbf{\mu_t}
\]
where $E(\bx) = \ell(\Phi(\bx), \Phi_0) + \lambda \mathcal{R}(\bx)$ is the objective function.
The vector $\mu_t$ is a weighed average of the last several gradients, with decaying factor $m=0.9$.
Learning proceeds a few hundred iterations with a fixed learning rate $\eta_t$ and is reduced tenfold, until convergence.

\paragraph{Computing derivatives.} Applying GD requires computing the derivatives of the loss function composed with the representation $\Phi(\bx)$.
While the squared Euclidean loss is smooth, this is not the case for the representation.
A key feature of CNNs is the ability of computing the derivatives of each computational layer, composing the latter in an overall derivative of the whole function using back-propagation.
Our translation of HOG and DSIFT into CNN allows us to apply the same technique to these computer vision representations too.

\section{Representations}\label{s:representations}

This section describes the image representations studied in the paper: DSIFT (Dense-SIFT), HOG, and reference  deep CNNs.
Furthermore, it shows how to implement DSIFT and HOG in a standard CNN framework in order to compute their derivatives.
Being able to compute derivatives is the only requirement imposed by the algorithm of Sect.~\ref{s:optimisation}.
Implementing DSIFT and HOG in a standard CNN framework makes derivative computation convenient.

\paragraph{CNN-A: deep networks.} As a reference deep network we consider the Caffe-Alex~\cite{jia13caffe} model (CNN-A), which closely reproduces the network by Krizhevsky \etal~\cite{krizhevsky12imagenet}.
This and many other similar networks alternate the following computational building blocks: linear convolution, ReLU gating, spatial max-pooling, and group normalisation.
Each such block takes as input a $d$-dimensional image and produces as output a $k$-dimensional one.
Blocks can additionally pad the image (with zeros for the convolutional blocks and with $-\infty$ for max pooling) or subsample the data.
The last several layers are deemed ``fully connected'' as the support of the linear filters coincides with the size of the image; however, they are equivalent to filtering layers in all other respects. Table~\ref{f:cnna} details the structure of CNN-A.
 
\paragraph{CNN-DSIFT and CNN-HOG.} This section shows how DSIFT~\cite{lowe99object,nowak06sampling} and HOG~\cite{dalal05histograms} can be implemented as CNNs.
This formalises the relation between CNNs and these standard representations.
It also makes derivative computation for these representations simple; for the inversion algorithm of Sect.~\ref{s:method}.
The DSIFT and HOG implementations in the VLFeat library~\cite{vedaldi07open} are used as numerical references. These are equivalent to Lowe's~\cite{lowe99object} SIFT and the DPM V5~HOG~\cite{lsvm-pami,voc-release5}. 

SIFT and HOG involve: computing and binning image gradients, pooling binned gradients into cell histograms, grouping cells into blocks, and normalising the blocks.
Denote by $\bg$ the gradient at a given pixel and consider binning this into one of $K$ orientations (where $K=8$ for SIFT and $K=18$ for HOG).
This can be obtained in two steps: directional filtering and gating.
The $k$-th directional filter is $G_k = u_{1k} G_x + u_{2k } G_y$ where
\[
\bu_k = \begin{bmatrix} \cos \frac{2\pi k}{K} \\ \sin \frac{2\pi k}{K} \end{bmatrix},
\quad
 G_x = \begin{bmatrix} 0 & 0 & 0 \\ -1 & 0 & 1 \\ 0 & 0 & 0 \end{bmatrix},
\quad
 G_y = G_x^\top.
\]
The output of a directional filter is the projection $\langle \bg, \bu_k \rangle$ of the gradient along direction $\bu_k$. 
A suitable gating function implements binning into a histogram element $h_k$. DSIFT uses bilinear orientation binning, given by
\[
  h_k= \|\bg\| 
  \max\left\{0, 1 - \frac{K}{2\pi} \cos^{-1} \frac{\langle \bg, \bu_k \rangle}{\|\bg\|} \right\},
\]
whereas HOG (in the DPM V5 variant) uses hard assignments $h_k = \|\bg\| \mathbf{1}\left[\langle \bg, \bu_k \rangle > \|\bg\| \cos\pi/K \right]$.
Filtering is a standard CNN operation but these binning functions are not.
While their implementation is simple, an interesting alternative is the approximated bilinear binning:
\begin{align*}
  h_k 
  &\approx \|\bg\| 
  \max\left\{0, \frac{1}{1-a} \frac{\langle \bg, \bu_k \rangle }{\|\bg\|} - \frac{a}{1-a}\right\}
  \\
  &\propto \max\left\{0, \langle \bg, \bu_k \rangle - a\|\bg\| \right\},
  \quad a = \cos 2\pi/K.
\end{align*}
The norm-dependent offset $\|\bg\|$ is still non-standard, but the ReLU operator is, which shows to which extent approximate binning can be achieved in typical CNNs.

The next step is to pool the binned gradients into cell histograms using bilinear spatial pooling, followed by extracting blocks of $2\times 2$ (HOG) or $4 \times 4$ (SIFT) cells.
Both such operations can be implemented by banks of linear filters.
Cell blocks are then $l^2$ normalised, which is a special case of the standard local response normalisation layer.
For HOG, blocks are further decomposed back into cells, which requires another filter bank.
Finally, the descriptor values are clamped from above by applying $y = \min\{x,0.2\}$ to each component, which can be reduced to a combination of linear and ReLU layers.

The conclusion is that approximations to DSIFT and HOG can be implemented with conventional CNN components plus the non-conventional gradient norm offset.
However, all the filters involved are much sparser and simpler than the generic 3D filters in learned CNNs. 
Nonetheless, in the rest of the paper we will use exact CNN equivalents of DSIFT and HOG, using modified or additional CNN components as needed.
\footnote{This requires addressing a few more subtleties. In DSIFT gradient contributions are usually weighted by a Gaussian centered at each descriptor (a $4 \times 4$ cell block); here we use the VLFeat approximation (\texttt{fast} option) of weighting cells rather than gradients, which can be incorporated in the block-forming filters. In UoCTTI HOG, cells contain both oriented and unoriented gradients (27 components in total) as well as 4 texture components. The latter are ignored for simplicity, while the unoriented gradients are obtained as average of the oriented ones in the block-forming filters. Curiously, in UoCTTI HOG the $l^2$ normalisation factor is computed considering only the unoriented gradient components in a block, but applied to all, which requires modifying the normalization operator. Finally, when blocks are decomposed back to cells, they are averaged rather than stacked as in the original Dalal-Triggs HOG, which can be implemented in the block-decomposition filters.} These CNNs are numerically indistinguishable from the VLFeat reference implementations, but, true to their CNN nature, allow computing the feature derivatives as required by the algorithm of Sect.~\ref{s:method}.

\newcommand{\puti}[2]
{%
\begin{tikzpicture}
\node[anchor=south east,inner sep=0] at (0,0) {#2};
\node[image label]{#1};
\end{tikzpicture}%
}

\begin{figure*}[ht!]
\newcommand{\tri}{trim={0 0 {.5\width} {.5\height}},clip}
\puti{(a) Orig.}{\includegraphics[width=0.166\textwidth]{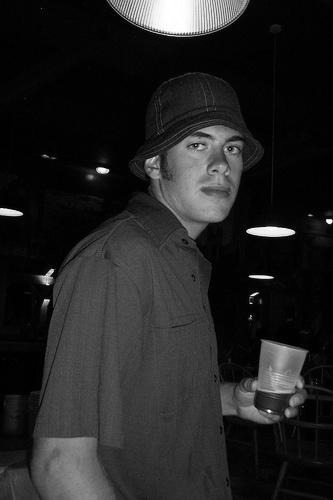}}%
\puti{(b) HOG}{\includegraphics[width=0.166\textwidth]{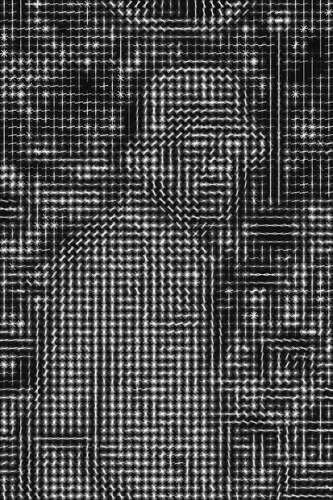}}%
\puti{(c) HOGgle~\cite{vondrick13hoggles:}}{\includegraphics[width=0.166\textwidth]{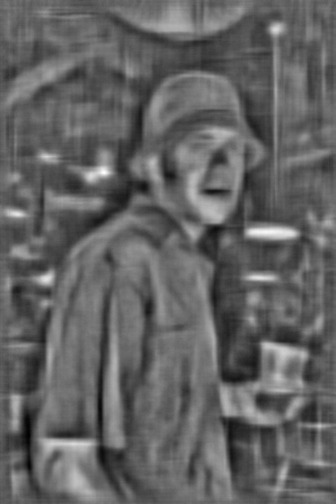}}%
\puti{(d) $\text{HOG}^{-1}$}{\includegraphics[width=0.166\textwidth]{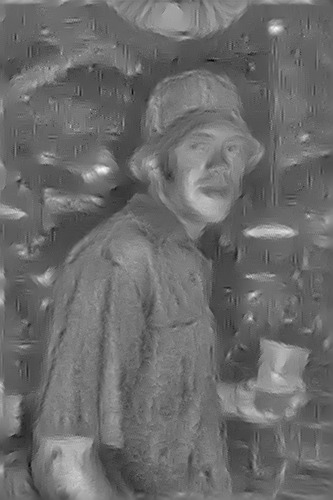}}%
\puti{(e) $\text{HOGb}^{-1}$}{\includegraphics[width=0.166\textwidth]{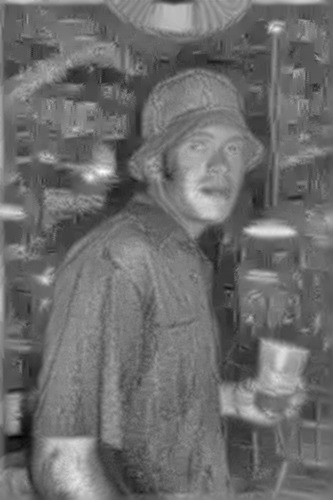}}%
\puti{(f) $\text{DSIFT}^{-1}$}{\includegraphics[width=0.166\textwidth]{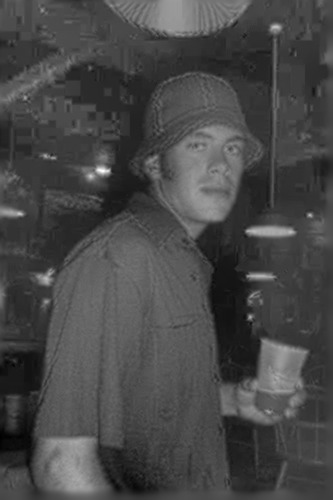}}
{\adjincludegraphics[width=0.166\textwidth,trim={150pt 300pt 80pt 100pt},clip]{figures/v26/bfhog-tv2/hoggle-orig-1/lInf-orig}}%
{\adjincludegraphics[width=0.166\textwidth,trim={150pt 300pt 80pt 100pt},clip]{figures/v26/hog-hoggle/hoggle-orig-1/lInf-hog}}%
{\adjincludegraphics[width=0.166\textwidth,trim={150pt 300pt 80pt 100pt},clip]{figures/v26/hog-hoggle/hoggle-orig-1/lInf-recon}}%
{\adjincludegraphics[width=0.166\textwidth,trim={150pt 300pt 80pt 100pt},clip]{figures/v26/bfhog-tv2/hoggle-orig-1/lInf-recon}}%
{\adjincludegraphics[width=0.166\textwidth,trim={150pt 300pt 80pt 100pt},clip]{figures/v26/bfhogb-tv2/hoggle-orig-1/lInf-recon}}%
{\adjincludegraphics[width=0.166\textwidth,trim={150pt 300pt 80pt 100pt},clip]{figures/v26/bfdsift-tv2/hoggle-orig-1/lInf-recon}}
\caption{Reconstruction quality of different representation inversion methods, applied to HOG and DSIFT. HOGb denotes HOG with bilinear orientation assignments. This image is best viewed on screen.}\label{f:hoggles}
\end{figure*}

Next we apply the algorithm from Sect.~\ref{s:method} on \textbf{CNN-A}, \textbf{CNN-DSIFT} and \textbf{CNN-HOG} to analyze our method.

\section{Experiments with shallow representations}\label{s:results-shallow}
\begin{table}
\centering
\begin{tabular}{|c|cc|c|c|}
\hline
descriptors & HOG & HOG & HOGb & DSIFT \\
method & HOGgle & our & our & our \\
\hline
error (\%) &$ 66.20$&$ 28.10$&$ 10.67$&$ 10.89$\\[-0.5em]
~&\tiny$\pm13.7$&\tiny$\pm 7.9$&\tiny$\pm 5.2$&\tiny$\pm 7.5$\\
\hline
\end{tabular}
\vspace{0.5em}
\caption{Average reconstruction error of different representation inversion methods, applied to HOG and DSIFT. HOGb denotes HOG with bilinear orientation assignments. The standard deviation shown is the standard deviation of the error and not the standard deviation of the mean error.}\label{t:hog-errors}
\end{table}

\begin{figure}
\hfill
{\adjincludegraphics[width=0.2\columnwidth,trim={180pt 300pt 80pt 140pt},clip]{figures/v26/bfhog-tv2/hoggle-orig-1/lInf-orig}}
\hfill
{\adjincludegraphics[width=0.2\columnwidth,trim={180pt 300pt 80pt 140pt},clip]{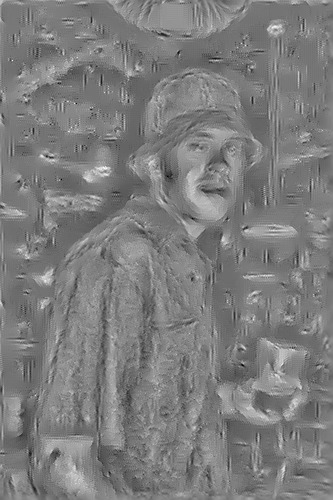}}
\hfill
{\adjincludegraphics[width=0.2\columnwidth,trim={180pt 300pt 80pt 140pt},clip]{figures/v26/bfhog-tv2/hoggle-orig-1/lInf-recon}}
\hfill
{\adjincludegraphics[width=0.2\columnwidth,trim={180pt 300pt 80pt 140pt},clip]{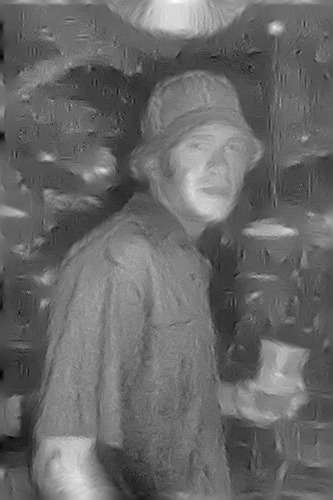}}
\hfill\mbox{}
\caption{Effect of $\TV$ regularization. The same inversion algorithm visualized in Fig.~\ref{f:hoggles}(d) is used with a smaller ($\lambda_{\TV} = 0.5$), comparable ($\lambda_{\TV} = 5.0$), and larger ($\lambda_{\TV} = 50$) regularisation coefficient.}\label{f:tv}
\end{figure}

This section evaluates the representation inversion method of Sect.~\ref{s:method} by applying it to HOG and DSIFT.
The analysis includes both a qualitative (Fig.~\ref{f:hoggles}) and quantitative (Table~\ref{t:hog-errors}) comparison with existing technique.
The quantitative evaluation reports a normalized reconstruction error $\|\Phi(\bx^*) - \Phi(\bx_i)\|_2/\mathrm{N}_{\Phi}$ averaged over $100$ images $\bx_i$ from the ILSVRC 2012 challenge~\cite{ILSVRCarxiv14} validation data (images 1 to 100). 
A normalization is essential to place the Euclidean distance in the context of the volume occupied by the features: if the features are close together, then even an Euclidean distance of $0.1$ is very large, 
but if the features are spread out, then even an Euclidean distance of $10^5$ may be very small.
We use $\mathrm{N}_{\Phi}$ to be the average pairwise euclidean distance between $\Phi(\bx_i)$'s across the 100 test images.

We fix the parameters in equation \ref{eq:obj} to $\lambda_\alpha = 2.16\times 10^{8}$, $\lambda_{\TV} = 5$, and $\beta = 2$.

The closest alternative to our method is HOGgle, a technique introduced by Vondrick~\etal~\cite{vondrick13hoggles:} for the visualisation of HOG features.
The HOGgle code is publicly available from the authors' website and is used throughout these experiments. 
Crucially, HOGgle is pre-trained to invert the UoCTTI implementation of HOG, which is numerically equivalent to CNN-HOG (Sect.~\ref{s:representations}), allowing for a direct comparison between algorithms.

Compared to our method, HOGgle is fast (2-3s vs 60s on the same CPU) but not very accurate, as it is apparent both qualitatively (Fig.~\ref{f:hoggles}.c vs d) and quantitatively (66\% vs 28\% reconstruction error, see Table.~\ref{t:hog-errors}).
Interestingly, \cite{vondrick13hoggles:} propose a direct optimisation method similar to~\eqref{e:objective}, but show that it does not perform better than HOGgle.
This demonstrates the importance of the choice of regulariser and the ability of computing the derivative of the representation.
The effect of the regularizer $\lambda_{\TV}$ is further analysed in Fig.~\ref{f:tv} (and later in Table~\ref{t:cnn-errors}): without this prior information, the reconstructions present a significant amount of discretization artifacts.

In terms of speed, an advantage of optimizing~\eqref{e:objective} is that it can be switched to use GPU code immediately given the underlying CNN framework; doing so results in a ten-fold speedup.
Furthermore the CNN-based implementation of HOG and DSIFT wastes significant resources using generic filtering code despite the particular nature of the filters in these two representations.
Hence we expect that an optimized implementation could be several times faster than this.

It is also apparent that different representations can be easier or harder to invert.
In particular, modifying HOG to use bilinear gradient orientation assignments as SIFT (Sect.~\ref{s:representations}) significantly reduces the reconstruction error (from 28\% down to 11\%) and improves the reconstruction quality (Fig.~\ref{f:hoggles}.e).
More impressive is DSIFT: it is quantitatively similar to HOG with bilinear orientations, but produces significantly more detailed images (Fig.~\ref{f:hoggles}.f).
Since HOG uses a finer quantisation of the gradient compared to SIFT but otherwise the same cell size and sampling, this result can be imputed to the heavier block-normalisation of HOG that evidently discards more image information than SIFT.

\section{Experiments with deep representations}\label{s:results-deep}
\begin{figure}
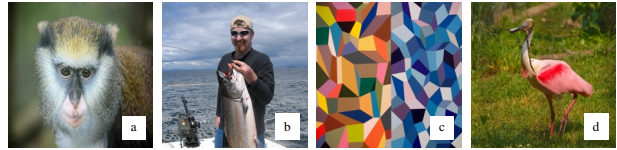

\newcommand{\putx}[2]{%
\puti{#1}{\noexpand{\adjincludegraphics[width=0.11\textwidth]{figures/v26/bfcnna1-group1/#2/l01-orig}}}}%
\hfill%
\putx{a}{ILSVRC2012_val_00000013}\hfill%
\putx{b}{stock_fish}\hfill%
\putx{c}{stock_abstract}\hfill%
\putx{d}{ILSVRC2012_val_00000043}\hfill\mbox{}
\caption{Test images for qualitative results.}\label{f:test-image}
\end{figure}

\begin{table*}
\setlength{\tabcolsep}{2pt}
\footnotesize
\centering
\begin{tabular}{|l|cccccccccccccccccccc|}
\hline
     layer&      1&      2&      3&      4&      5&      6&      7&      8&      9&     10&     11&     12&     13&     14&     15&     16&     17&     18&     19&     20 \\
\hline
      name& conv1 &  relu1& mpool1& norm1 & conv2 & relu2 & mpool2& norm2 & conv3 & relu3 & conv4 & relu4 & conv5 & relu5 & mpool5 &  fc6 & relu6 &   fc7 &  relu7&    fc8 \\
      type&    cnv&   relu&  mpool&    nrm&    cnv&   relu&  mpool&    nrm&    cnv&   relu&    cnv&   relu&    cnv&   relu&  mpool&    cnv&   relu&    cnv&   relu&    cnv \\
 channels&     96&     96&     96&     96&    256&    256&    256&    256&    384&    384&    384&    384&    256&    256&    256&   4096&   4096&   4096&   4096&   1000 \\
\hline
rec. field&     11&     11&     19&     19&     51&     51&     67&     67&     99&     99&    131&    131&    163&    163&    195&    355&    355&    355&    355&    355    \\
\hline
\end{tabular}
\vspace{0.5em}
\caption{{\bf CNN-A structure.} The table specifies the structure of CNN-A along with receptive field size of each neuron. The filters in layers from 16 to 20 operate as ``fully connected'': given the standard image input size of $227 \times 227$ pixels, their support covers the whole image. Note also that their receptive field is larger than 227 pixels, but can be contained in the image domain due to padding.}\label{f:cnna}
\end{table*}

\begin{figure*}
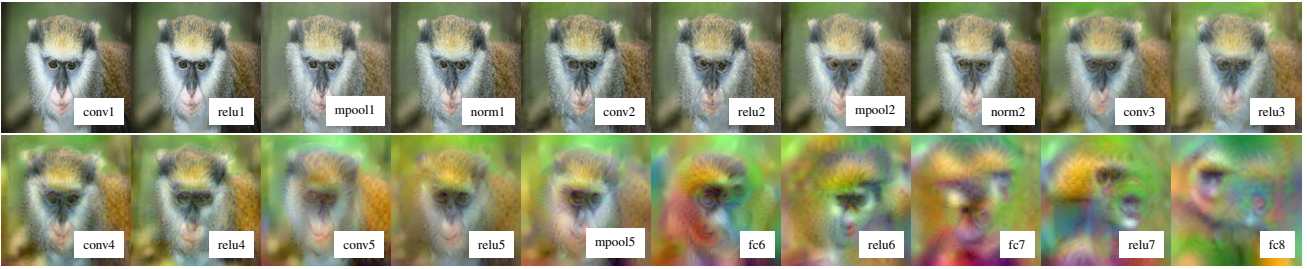

\centering
\newcommand{\putx}[3]{%
\puti{#1}{\noexpand{\adjincludegraphics[width=0.099\textwidth]{figures/v26/#2/ILSVRC2012_val_00000013/l#3-recon}}}}%
\putx{conv1}{bfcnna1}{01}%
\putx{relu1}{bfcnna1}{02}%
\putx{mpool1}{bfcnna1}{03}%
\putx{norm1}{bfcnna1}{04}%
\putx{conv2}{bfcnna1}{05}%
\putx{relu2}{bfcnna1}{06}%
\putx{mpool2}{bfcnna2}{07}%
\putx{norm2}{bfcnna2}{08}%
\putx{conv3}{bfcnna2}{09}%
\putx{relu3}{bfcnna2}{10}
\putx{conv4}{bfcnna2}{11}%
\putx{relu4}{bfcnna2}{12}%
\putx{conv5}{bfcnna3}{13}%
\putx{relu5}{bfcnna3}{14}%
\putx{mpool5}{bfcnna3}{15}%
\putx{fc6}{bfcnna3}{16}%
\putx{relu6}{bfcnna3}{17}%
\putx{fc7}{bfcnna3}{18}%
\putx{relu7}{bfcnna3}{19}%
\putx{fc8}{bfcnna3}{20}
\caption{{\bf CNN reconstruction.} Reconstruction of the image of Fig.~\ref{f:test-image}.a from each layer of CNN-A. To generate these results, the regularization coefficient for each layer is chosen to match the highlighted rows in table~\ref{t:cnn-errors}. This figure is best viewed in color/screen.}\label{f:cnn-layers}
\end{figure*}

\begin{table*}
\setlength{\tabcolsep}{1pt}
\begin{tabular}{|c||cccccccccccccccccccc|}
\hline
          $\lambda_{\TV}$            & 1            & 2            & 3            & 4            & 5            & 6            & 7            & 8            & 9           & 10           & 11           & 12           & 13           & 14           & 15           & 16           & 17           & 18           & 19           & 20             \\
                                   & conv1        & relu1        & pool1        & norm1        & conv2        & relu2        & pool2        & norm2        & conv3        & relu3        & conv4        & relu4        & conv5        & relu5        & pool5          & fc6        & relu6          & fc7        & relu7          & fc8             \\
         \hline
          $\lambda_1$       & $\mathbf{10.0}$       & $\mathbf{11.3}$       & $\mathbf{21.9}$       & $\mathbf{20.3}$       & $\mathbf{12.4}$       & $\mathbf{12.9}$       & $15.5$       & $15.9$       & $14.5$       & $16.5$       & $14.9$       & $13.8$       & $12.6$       & $15.6$       & $16.6$       & $12.4$       & $15.8$       & $12.8$       & $10.5$        & $5.3$     \\[-0.5em]
            &\tiny$\pm5.0$ &\tiny$\pm5.5$ &\tiny$\pm9.2$ &\tiny$\pm5.0$ &\tiny$\pm3.1$ &\tiny$\pm5.3$ &\tiny$\pm4.7$ &\tiny$\pm4.6$ &\tiny$\pm4.7$ &\tiny$\pm5.3$ &\tiny$\pm3.8$ &\tiny$\pm3.8$ &\tiny$\pm2.8$ &\tiny$\pm5.1$ &\tiny$\pm4.6$ &\tiny$\pm3.5$ &\tiny$\pm4.5$ &\tiny$\pm6.4$ &\tiny$\pm1.9$ &\tiny$\pm1.1$             \\
          $\lambda_2$       & $20.2$       & $22.4$       & $30.3$       & $28.2$       & $20.0$       & $17.4$       & $\mathbf{18.2}$       & $\mathbf{18.4}$       & $\mathbf{14.4}$       & $\mathbf{15.1}$       & $\mathbf{13.3}$       & $\mathbf{14.0}$       & $15.4$       & $13.9$       & $15.5$       & $14.2$       & $13.7$       & $15.4$       & $10.8$        & $5.9$     \\[-0.5em]
            &\tiny$\pm9.3$&\tiny$\pm10.3$&\tiny$\pm13.6$ &\tiny$\pm7.6$ &\tiny$\pm4.9$ &\tiny$\pm5.0$ &\tiny$\pm5.5$ &\tiny$\pm5.0$ &\tiny$\pm3.6$ &\tiny$\pm3.3$ &\tiny$\pm2.6$ &\tiny$\pm2.8$ &\tiny$\pm2.7$ &\tiny$\pm3.2$ &\tiny$\pm3.5$ &\tiny$\pm3.7$ &\tiny$\pm3.1$&\tiny$\pm10.3$ &\tiny$\pm1.6$ &\tiny$\pm0.9$             \\
          $\lambda_3$       & $40.8$       & $45.2$       & $54.1$       & $48.1$       & $39.7$       & $32.8$       & $32.7$       & $32.4$       & $25.6$       & $26.9$       & $23.3$       & $23.9$       & $\mathbf{25.7}$       & $\mathbf{20.1}$       & $\mathbf{19.0}$       & $\mathbf{18.6}$       & $\mathbf{18.7}$       & $\mathbf{17.1}$       & $\mathbf{15.5}$        & $\mathbf{8.5}$     \\[-0.5em]
           &\tiny$\pm17.0$&\tiny$\pm18.7$&\tiny$\pm22.7$&\tiny$\pm11.8$ &\tiny$\pm9.1$ &\tiny$\pm7.7$ &\tiny$\pm8.0$ &\tiny$\pm7.0$ &\tiny$\pm5.6$ &\tiny$\pm5.2$ &\tiny$\pm4.1$ &\tiny$\pm4.6$ &\tiny$\pm4.3$ &\tiny$\pm4.3$ &\tiny$\pm4.3$ &\tiny$\pm4.9$ &\tiny$\pm3.8$ &\tiny$\pm3.4$ &\tiny$\pm2.1$ &\tiny$\pm1.3$             \\
\hline
\end{tabular}
\vspace{0.1em}
\caption{{\bf Inversion error for CNN-A.} Average inversion percentage error (normalized) for all the layers of CNN-A and various amounts of $\TV$ regularisation: $\lambda_1=0.5$, $\lambda_2=10\lambda_1$ and $\lambda_3=100\lambda_1$. In bold face are the error values corresponding to the regularizer that works best both qualitatively and quantitatively. The deviations specified in this table are the standard deviations of the errors and not the standard deviations of the mean error value.}\label{t:cnn-errors}
\end{table*}

\begin{figure*}[ht!]
\newcommand{\putx}[3]{%
\puti{#1}{\noexpand{\adjincludegraphics[width=0.23\textwidth,trim={0 0 {0.3333\width} 0},clip]{figures/v26/bfcnna#2-repeats/\which/l#3-recon}}}}%
\begin{center}
\newcommand{\which}{stock_abstract}
\putx{pool5}{1}{15} %
\putx{relu6}{2}{17} %
\putx{relu7}{3}{19} %
\putx{fc8}{3}{20}
\renewcommand{\which}{ILSVRC2012_val_00000043}
\putx{pool5}{1}{15} %
\putx{relu6}{2}{17} %
\putx{relu7}{3}{19} %
\putx{fc8}{3}{20} 
\end{center}
\vspace{-1em}
\caption{{\bf CNN invariances.} Multiple reconstructions of the images of Fig.~\ref{f:test-image}.c--d from different deep codes obtained from CNN-A. This figure is best seen in colour/screen.}\label{f:cnn-invariance}
\end{figure*}

\begin{figure}[ht!]
\hfill
{\adjincludegraphics[width=0.15\textwidth]{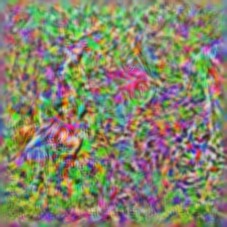}}
\hfill
{\adjincludegraphics[width=0.15\textwidth]{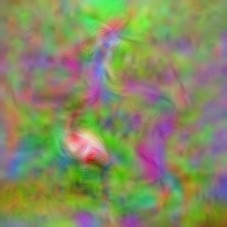}}
\hfill
{\adjincludegraphics[width=0.15\textwidth]{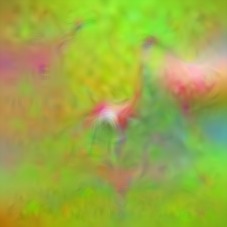}}
\hfill\mbox{}
\caption{Effect of $\TV$ regularization on CNNs. Inversions of the last layers of CNN-A for Fig.~\ref{f:test-image}.d with a progressively larger regulariser $\lambda_{\TV}$. This image is best viewed in color/screen.}\label{f:cnn-tv}
\end{figure}

\begin{figure*}[ht!]
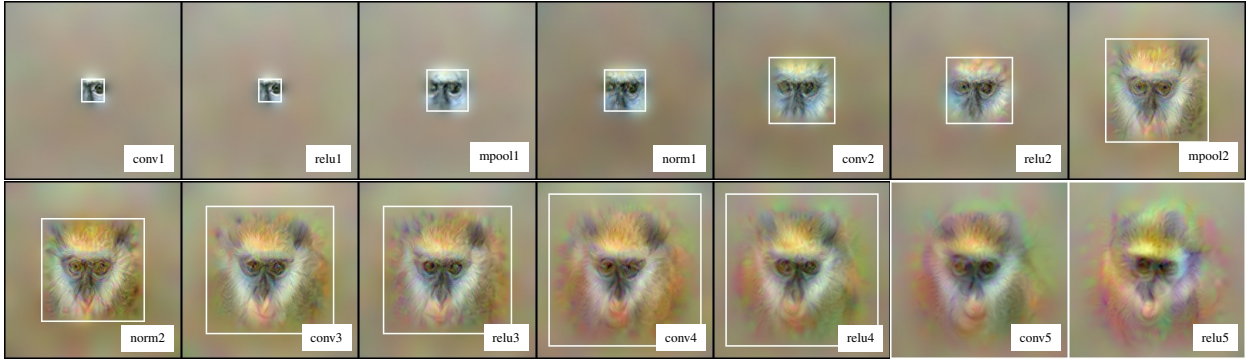

\centering
\newcommand{\putx}[3]{%
\puti{#1}{\noexpand{\adjincludegraphics[width=0.135\textwidth]{figures/v26/bfcnna#2-neigh5/ILSVRC2012_val_00000013/l#3-recon-fovoverlaid}}}}%
\putx{conv1}{3}{01}%
\putx{relu1}{3}{02}%
\putx{mpool1}{3}{03}%
\putx{norm1}{3}{04}%
\putx{conv2}{3}{05}%
\putx{relu2}{3}{06}%
\putx{mpool2}{3}{07}
\putx{norm2}{3}{08}%
\putx{conv3}{3}{09}%
\putx{relu3}{3}{10}%
\putx{conv4}{3}{11}%
\putx{relu4}{3}{12}%
\putx{conv5}{3}{13}%
\putx{relu5}{3}{14}%
\caption{{\bf CNN receptive field.} Reconstructions of the image of Fig.~\ref{f:test-image}.a from the central $5\times 5$ neuron fields at different depths of CNN-A. The white box marks the field of view of the $5\times 5$ neuron field. The field of view is the entire image for conv5 and relu5.}\label{f:cnn-neigh}
\end{figure*}

\begin{figure*}[ht!]
\centering
\newcommand{\putx}[2]{%
\puti{#1}{\noexpand{\includegraphics[width=0.160\textwidth]{figures/v26/bfcnna1-group2/stock_abstract/l#2-recon}}}}%
\putx{conv1-grp1}{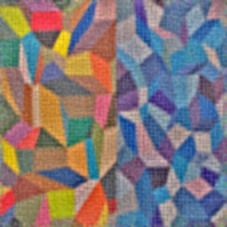}%
\putx{norm1-grp1}{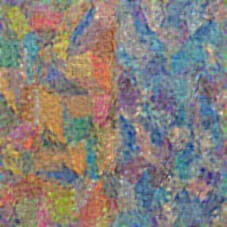}%
\putx{norm2-grp1}{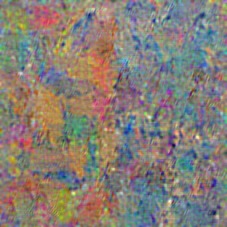}%
\renewcommand{\putx}[2]{%
\puti{#1}{\noexpand{\includegraphics[width=0.160\textwidth]{figures/v26/bfcnna1-group2/stock_fish/l#2-recon}}}}%
\putx{conv1-grp1}{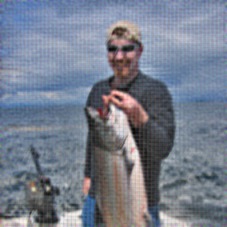}%
\putx{norm1-grp1}{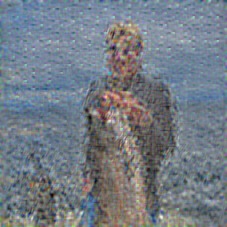}%
\putx{norm2-grp1}{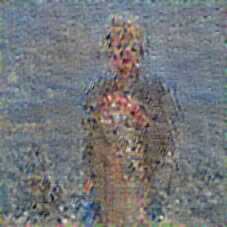}
\renewcommand{\putx}[2]{%
\puti{#1}{\noexpand{\includegraphics[width=0.160\textwidth]{figures/v26/bfcnna1-group1/stock_abstract/l#2-recon}}}}%
\putx{conv1-grp2}{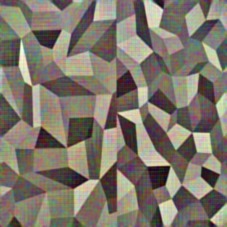}%
\putx{norm1-grp2}{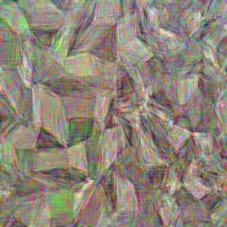}%
\putx{norm2-grp2}{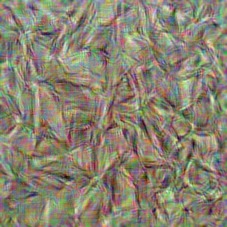}%
\renewcommand{\putx}[2]{%
\puti{#1}{\noexpand{\includegraphics[width=0.160\textwidth]{figures/v26/bfcnna1-group1/stock_fish/l#2-recon}}}}%
\putx{conv1-grp2}{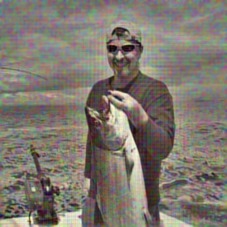}%
\putx{norm1-grp2}{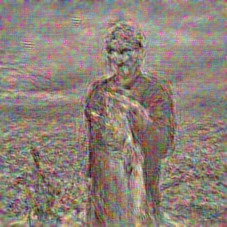}%
\putx{norm2-grp2}{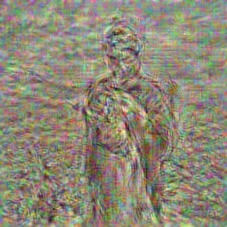}%
\caption{{\bf CNN neural streams.} Reconstructions of the images of Fig.~\ref{f:test-image}.c-b from either of the two neural streams of CNN-A. This figure is best seen in colour/screen.}\label{f:cnn-streams}
\end{figure*}
\vspace{-0.5em}
This section evaluates the inversion method applied to CNN-A described in Sect.~\ref{s:representations}.
Compared to CNN-HOG and CNN-DSIFT, this network is significantly larger and deeper.
It seems therefore that the inversion problem should be considerably harder. 
Also, CNN-A is not handcrafted but learned from 1.2M images of the ImageNet ILSVRC 2012 data~\cite{ILSVRCarxiv14}.

The algorithm of Sect.~\ref{s:optimisation} is used to invert the code obtained from each individual CNN layer for 100 ILSVRC validation images (these were not used to train the CNN-A model~\citep{krizhevsky12imagenet}). 
Similar to Sect.~\ref{s:results-shallow}, the normalized inversion error is computed and reported in Table~\ref{t:cnn-errors}.
The experiment is repeated by fixing $\lambda_\alpha$ to a fixed value of $2.16\times10^{8}$ and gradually increasing $\lambda_{\TV}$ ten-folds, starting from a relatively small value $\lambda_1 = 0.5$.
The ImageNet ILSVRC mean image is added back to the reconstruction before visualisation as this is subtracted when training the network. 
Somewhat surprisingly, the quantitative results show that CNNs are, in fact, not much harder to invert than HOG. 
The error rarely exceeds $20\%$, which is comparable to the accuracy of HOG (Sect.~\ref{s:results-shallow}).
The last layer is in particular easy to invert with an average error of $8.5\%$.

We choose the regularizer coefficients for each representation/layer based on a quantitative and qualitative study of the reconstruction.
We pick $\lambda_1 = 0.5$ for layers 1-6, $\lambda_2 = 5.0$ for layers 7-12 and $\lambda_3 = 50$ for layers 13-20.
The error value corresponding to these parameters is marked in bold face in table \ref{t:cnn-errors}.
Increasing $\lambda_{\TV}$ causes a deterioration for the first layers, but for the latter layers it helps recover a more visually interpretable reconstruction.
Though this parameter can be tuned by cross validation on the normalized reconstruction error, a selection based on qualitative analysis is preferred because the method should yield images that are visually meaningful. 

Qualitatively, Fig.~\ref{f:cnn-layers} illustrates the reconstruction for a test image from each layer of CNN-A.
The progression is remarkable.
The first few layers are essentially an invertible code of the image.
All the convolutional layers maintain a photographically faithful representation of the image, although with increasing fuzziness.
The 4,096-dimensional fully connected layers are perhaps more interesting, as they invert back to a \emph{composition of parts similar but not identical to the ones found in the original image}.
Going from relu7 to fc8 reduces the dimensionality further to just 1,000; nevertheless some of these visual elements can still be identified.
Similar effects can be observed in the reconstructions in~Fig.~\ref{f:cnn-invariance}.
This figure includes also the reconstruction of an abstract pattern, which is not included in any of the ImageNet classes; still, all CNN codes capture distinctive visual features of the original pattern, clearly indicating that even very deep layers capture visual information.

Next, Fig.~\ref{f:cnn-invariance} examines the invariance captured by the CNN model by considering multiple reconstructions out of each deep layer.
A careful examination of these images reveals that the codes capture progressively larger deformations of the object.
In the ``flamingo'' reconstruction, in particular, relu7 and fc8 invert back to multiple copies of the object/parts at different positions and scales.

Note that all these and the original images are nearly indistinguishable from the viewpoint of the CNN model; it is therefore interesting to note the lack of detail in the deepest reconstructions, showing that the network captures just a sketch of the objects, which evidently suffices for classification.
Considerably lowering the regulariser parameter still yields very accurate inversions, but this time with barely any resemblance to a natural image.
This confirms that CNNs have strong non-natural confounders.

We now examine reconstructions obtained from subset of neural responses in different CNN layers.
Fig.~\ref{f:cnn-neigh} explores the \emph{locality} of the codes by reconstructing a central $5\times 5$ patch of features in each layer.
The regulariser encourages portions of the image that do not contribute to the neural responses to be switched off.
The locality of the features is obvious in the figure; what is less obvious is that the effective receptive field of the neurons is in some cases significantly smaller than the theoretical one - shown as a white box in the image.

Finally, Fig.~\ref{f:cnn-streams} reconstructs images from a subset of feature channels.
CNN-A contains in fact two subsets of feature channels which are independent for the first several layers (up to norm2)~\cite{krizhevsky12imagenet}.
Reconstructing from each subset individually, clearly shows that one group is tuned towards low-frequency colour information  whereas the second one is tuned to towards high-frequency luminance components.
Remarkably, this behaviour emerges naturally in the learned network without any mechanism directly encouraging this pattern.

\begin{figure}[h!]
\centering
\newcommand{\putx}[4]{%
\noexpand{\adjincludegraphics[width=0.092\textwidth]{figures/v26/#1/ILSVRC2012_val_000000#2/l#3-#4}}}%
\putx{bfcnna3}{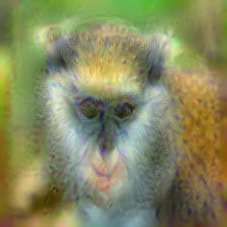}{01}{orig}%
\putx{bfcnna3}{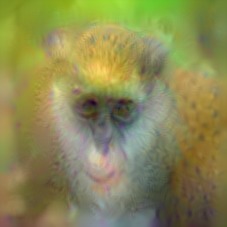}{01}{orig}%
\putx{bfcnna3}{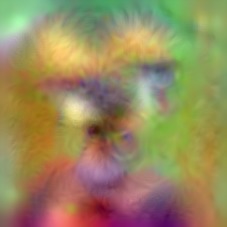}{01}{orig}%
\putx{bfcnna3}{23}{01}{orig}%
\putx{bfcnna3}{33}{01}{orig}
\putx{bfcnna3}{11}{15}{recon}%
\putx{bfcnna3}{14}{15}{recon}%
\putx{bfcnna3}{18}{15}{recon}%
\putx{bfcnna3}{23}{15}{recon}%
\putx{bfcnna3}{33}{15}{recon}
\caption{{\bf Diversity in the CNN model.} mpool5 reconstructions show that the network retains rich information even at such deep levels. This figure is best viewed in color/screen (zoom in).}\label{f:cnn-diversity}
\end{figure}

\section{Summary}\label{s:summary}

This paper proposed an optimisation method to invert shallow and deep representations based on optimizing an objective function with gradient descent.
Compared to alternatives, a key difference is the use of image priors such as the $\TV$ norm that can recover the low-level image statistics removed by the representation.
This tool performs better than alternative reconstruction methods for HOG.
Applied to CNNs, the visualisations shed light on the information represented at each layer.
In particular, it is clear that a progressively more invariant and abstract notion of the image content is formed in the network.

In the future, we shall experiment with more expressive natural image priors and analyze the effect of network hyper-parameters on the reconstructions.
We shall extract subsets of neurons that encode object parts and try to establish sub-networks that capture different details of the image.

\footnotesize
\bibliographystyle{ieee}
\bibliography{local}
\end{document}